\documentclass[preprints,article,accept,moreauthors,pdftex]{Definitions/mdpi}

\usepackage{lineno}
\usepackage{graphicx}
\usepackage{times}
\usepackage{epsfig}
\usepackage{amsmath}
\usepackage{subcaption}
\usepackage{booktabs}
\usepackage{color}
\usepackage{longtable}
\usepackage{ulem}
\usepackage{setspace}
\usepackage{url} 
\usepackage{colortbl}

\newcommand{\etal}{\textit{et al}., }
\newcommand{\ie}{\textit{i}.\textit{e}., }
\newcommand{\eg}{\textit{e}.\textit{g}., }

\firstpage{1} 
\pubvolume{xx}
\issuenum{1}
\articlenumber{5}
\pubyear{2019}
\copyrightyear{2019}
\history{Received: date; Accepted: date; Published: date}

\Title{Automatic Hierarchical Classification of Kelps using Deep Residual Features}

\Author{Ammar Mahmood $^{1}$*\orcidA{}, Ana Giraldo Ospina $^{2}$, Mohammed Bennamoun $^{1}$, Senjian An $^{3}$, Ferdous Sohel $^{4}$, Farid Boussaid $^{5}$, Renae Hovey $^{2}$, Robert B. Fisher $^{6}$ and Gary A. Kendrick $^{2}$}

\AuthorNames{Ammar Mahmood, Ana Giraldo Ospina, Mohammed Bennamoun, Senjian An, Ferdous Sohel, Farid Boussaid, Renae Hovey, Robert B. Fisher and Gary A. Kendrick}

\address{%
$^{1}$ \quad Computer Science and Software Engineering, The University of Western Australia, WA 6009, Australia\\
$^{2}$ \quad School of Biological Sciences and Oceans Institute, The University of Western Australia, WA 6009, Australia\\
$^{3}$ \quad School of Electrical Engineering, Computing and Mathematical Sciences, Curtin University, WA 6845, Australia\\
$^{4}$ \quad College of Science, Health, Engineering and Education Murdoch University, WA 6150, Australia\\
$^{5}$ \quad Electrical, Electronic and Computer Engineering, The University of Western Australia, WA 6009, Australia\\
$^{6}$ \quad School of Informatics, University of Edinburgh, Edinburgh EH8 9YL, UK}

\corres{Correspondence: ammarmahmood@live.com}

\abstract{Across the globe, remote image data is rapidly being collected for the assessment of benthic communities from shallow to extremely deep waters on continental slopes to the abyssal seas. Exploiting this data is presently limited by the time it takes for experts to identify organisms found in these images.  With this limitation in mind, a large effort has been made globally to introduce automation and machine learning algorithms to accelerate both classification and assessment of marine benthic biota. One major issue lies with organisms that move with swell and currents, like kelps.  This paper presents an automatic hierarchical classification method (local binary classification as opposed to the conventional flat classification) to classify kelps in images collected by autonomous underwater vehicles. The proposed kelp classification approach exploits learned feature representations extracted from deep residual networks. We show that these generic features outperform the traditional off-the-shelf CNN features and the conventional hand-crafted features. Experiments also demonstrate that the hierarchical classification method outperforms the traditional parallel multi-class classifications by a significant margin (90.0\% vs 57.6\% and 77.2\% vs 59.0\%) on Benthoz15 and Rottnest datasets respectively.  Furthermore, we compare different hierarchical classification approaches and experimentally show that the sibling hierarchical training approach outperforms the inclusive hierarchical approach by a significant margin. We also report an application of our proposed method to study the change in kelp cover over time for annually repeated AUV surveys.}


\begin{document}

\section{Introduction}
\label{S:1}

Kelp forests support diverse and productive ecological communities throughout temperate and arctic regions worldwide. Environmental anomalies such as cyclones, storms, marine heat waves and climate change have a detrimental effect on benthic marine life including kelps \cite{doney2012climate}. Significant declines in kelp bed have been observed around the globe in the past decades, with the main drivers identified as eutrophication and climate change related environmental stressors. For instance, large-scale disappearance of kelp was observed in 2002 in the southern coast of Norway  \cite{moy2012large}. In Spain, large scale reductions in two main species of kelp have also been observed since 1980’s  \cite{fernandez2011retreat}. 

Similarly, kelp populations in Australia have decreased as a consequence of climate change driven environmental stressors. In the east coast of Tasmania, the coverage of giant kelp \textit{Macrocystis pyrifera} in the present decade is around 9\% of the coverage in the 1940’s  \cite{johnson2011climate}. This decline is consistent with the intrusion of warmer, nutrient poor water from the East Australian Current, which now extends 350 km further south than in the 1940’s  \cite{ridgway2007long}.  
Wernberg \etal \cite{wernberg2016climate} reported a rapid climate-driven transition of kelp forests to seaweed turfs in the Australian temperate reef communities with kelp forests showing a 100km poleward contraction from their pre-heatwave distribution on the Western Australia coast. This trend is alarming for the numerous endemic species that rely on kelp forests for support. Loss of kelp forests is also a major threat for Australia's fishing and tourism industries, which generate more than 10 billion Australian dollars per annum \cite{bennett2016great}. There is thus a pressing and immediate need for monitoring  programs to document changes in kelp dominated habitats  along coastlines worldwide and especially in temperate Australia.

Autonomous underwater vehicles (AUVs) are emerging as highly effective tools for monitoring changes in benthic marine environments, because  \textbf{(i)} they can autonomously conduct non-destructive sampling in remote marine habitats; \textbf{(ii)} they can repeatedly survey the same spatial region to detect change over time; and \textbf{(iii)} they are fitted with a range of instrumentation to acquire both physical and biological data. AUVs have been used to monitor the marine benthos across temperate and tropical environments in Australia \cite{williams2012monitoring}, \cite{smale2012regional}; to survey invasive pest species \cite{barrett2010autonomous}; to document rapid loss of corals associated with warming events \cite{smale2012regional}, \cite{bridge2014variable}; to describe benthic community structure at depths greater than 1000 m \cite{sherman2009deep}; and assess environmental impacts of the Deepwater Horizon oil spill \cite{camilli2010tracking}. In a large-scale study of deep waters, the distribution patterns of kelp forests were investigated to provide useful insights on the effect of environmental changes on the kelp population \cite{marzinelli2015large}. The survey took an extremely long time to complete as marine biologists had to manually classify images and to identify kelp from imagery. 

AUV driven monitoring can generate large quantities of imagery. For example, an AUV deployed in Western Australia collected more than 15,000 stereo image pairs each day and was deployed  between 10 and 12 days each year \cite{smale2012regional}. Manual analysis of such a large number of images per deployment (150,000 to 200,000 stereo image pairs) takes a significant amount of time and effort and is the major bottleneck in data acquisition from AUV surveys. In order to promptly identify changes in benthic species, especially dominant habitat formers (such as kelps and corals), it is necessary to match image-analysis time to surveying time so data can be analyzed rapidly and identification of change patterns can be accomplished. Automatic classification is critical to speed up image analysis and consequently automatic classification of benthic species has raised interest in ecologists and computer scientists (such as \cite{marcos2005classification, denuelle2010kelp, bewley2012automated, beijbom2012automated, mahmood2016coral}). Nonetheless, automated classification of AUV collected imagery is challenging because images are captured in dynamic shallow water with little to no control on lighting and significant variations in what is visible and how it is perceived.

\begin{figure}[]
\begin{center}
   \includegraphics[width=1\linewidth,height=1.8\linewidth, keepaspectratio]{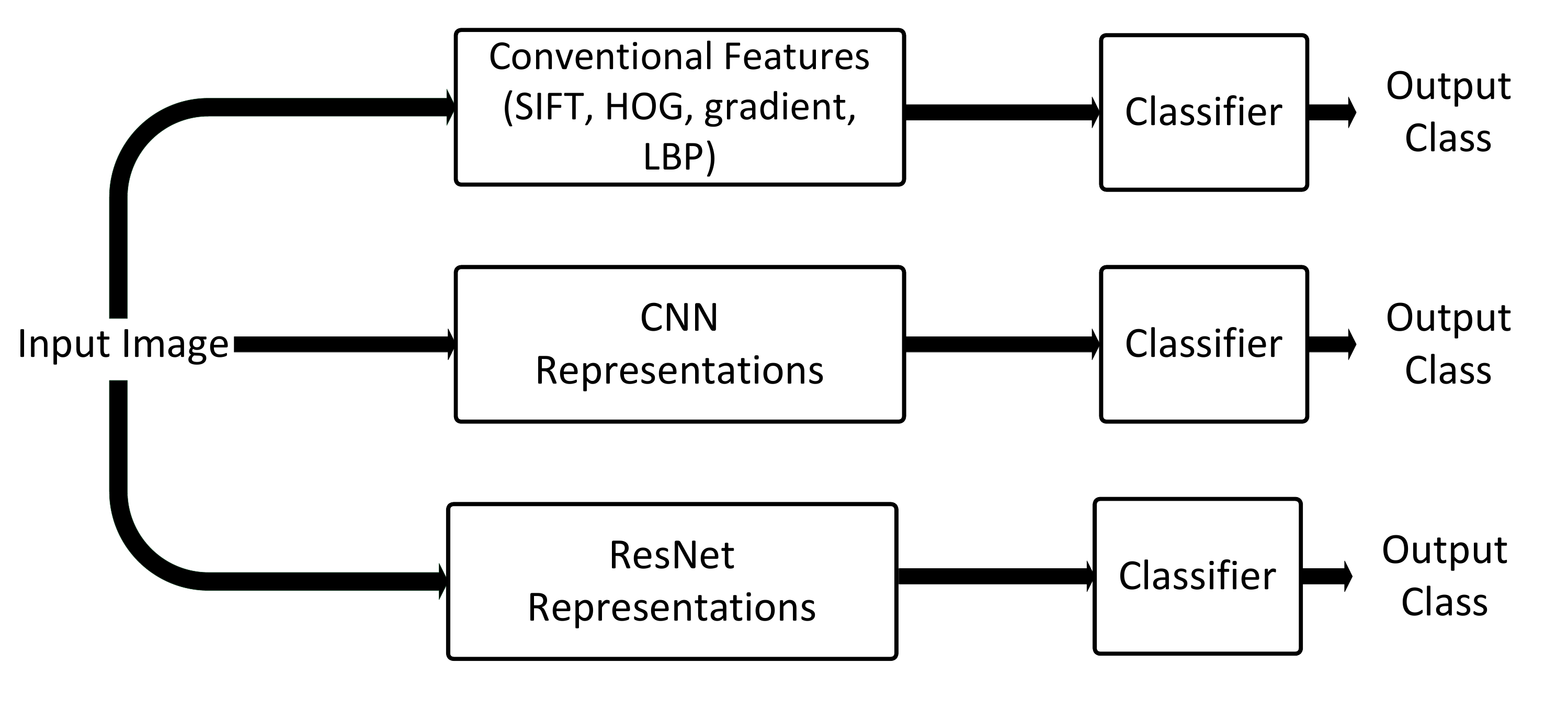}
\end{center}
  \caption{ \small{Evolution of classification pipelines (the most recent at the bottom).  Off-the-shelf deep residual features have the potential to replace the previous classification pipelines and improve performance for benthic marine image classification tasks. (SIFT: scale invariant feature transform, HOG: histograms of gradient, LBP: local binary patterns, CNN: convolutional neural networks, ResNet: residual networks.)} }

\label{fig:feat}
\end{figure}

In this paper, we tackle the challenge of automatically annotating underwater imagery for the presence of kelp to detect changes in the coverage of  Australian kelp forests. The common practice is to study the distribution and density of benthic species, which involves manually annotating a  smaller dataset and then extrapolating these results to make inferences about the sites under study. Automating the process of determining kelp coverage will significantly decrease image processing times and will allow for large scale analysis of  datasets and for early identification of changes in kelp cover.
To automate this process, it is paramount to select appropriate features. In computer vision tasks, the general trend has shifted from conventional hand-crafted features to off-the-shelf deep features \cite{razavian2014cnn}. Hand-crafted features which usually encode one aspect of data (\ie color, shape or texture)  were a popular choice as image representations for benthic marine species recognition tasks in the works of \cite{marcos2005classification, beijbom2012automated, stokes2009automated, pizarro2008towards}. Moreover, given that hand-crafted features are designed specifically for a current task at hand, they generally do not perform well when applied on a different task. Recently, Convolutional Neural Networks (CNNs) and features extracted from pre-trained CNNs have become the preferred choice for benthic marine image classification tasks, \eg \cite{mahmood2016coral, beijbom2016improving, mahmood2016automatic, mahmood2018deep}. These off-the-shelf features are image representations learned by a deep network trained on a larger dataset such as ImageNet. Off-the-shelf CNN features are generic and have shown better performance as compared to hand-crafted features on a variety of image recognition tasks \cite{razavian2014cnn}. In this paper, we propose to apply image representations extracted from deep residual networks (ResNets) to further improve the automatic annotation of benthic species. Besides better performance, one big advantage of
ResNets is their faster training time and ease of optimization. Figure \ref{fig:feat} depicts the evolution of classification pipelines for automatic benthic marine species annotation. \\ 

The main motivation for using ResNet as a base network to extract features for kelp classification is its superior performance over previous deep networks \cite{he2015deep}. Moreover, the feature extraction is  fast due to the low computational complexity of ResNets and the reduced number of floating point operations (FLOPs). Also, the feature extracted from ResNet is 2048-dimensional, which is half of the traditional 4096-dimensional feature vector of previous networks such as VGG16 \cite{simonyan2014very}. These compact features result in reduced memory requirements for storing the features of large benthic marine datasets. \\

The main contributions of this paper are:

\begin{enumerate}

\item  The first application of deep learning for automated kelp coverage analysis.
\item  A supervised kelp image classification method based on features extracted from deep residual networks, termed as Deep Residual Features (DRF).
\item A comparison of the classification performance of the DRF with the widely used off-the-shelf CNN features for automatic annotation of kelps. 
\item Experiments demonstrating DRF’s  superior  classification accuracy compared to previous methods for kelp classification. 
\item   We compare hierarchical image classification with multi-class image classification and report the accuracies and mean f1-scores for two large datasets. 
\item An application of our proposed method to automatically analyze kelp coverage across five regions of Rottnest Island in Western Australia.
\item We demonstrate the performance of the proposed kelp coverage analysis technique using ground truth data provided by marine experts and show a high correlation with previously conducted manual surveys.
\end{enumerate}

The paper is organized as follows.  In Section \ref{S:2}, we will briefly review related work. In Section \ref{S:3}, we present our proposed approach and explain the features extracted from deep networks. We then report the experimental results and kelp coverage analysis. In Section \ref{S:4}, we discuss the next steps required to implement our proposed method to a platform to rapidly analyze  benthic images. Section \ref{S:5} concludes this paper. 

\section{Related Work}
\label{S:2}

\subsection{Kelp Classification}

Previous studies on automatic classification and segmentation of kelps in benthic marine imagery were based on hand-crafted features (Table \ref{table:methods}). To the best of our knowledge, deep networks or features extracted from deep networks have not yet been applied to solve this problem. Here we briefly summarize  a few of the prominent studies focused on automating  kelp identification. 

Denuelle and Dunbabin \cite{denuelle2010kelp} utilized a technique that employed  generation of kelp probability maps using Haralick texture features across an entire image. They reported that supervised and unsupervised segmentation yielded similar results. Color imbalance resulted in a significant number of false positives thus implying that the images collected must be diversified to cater for the various possible underwater lighting  and visibility conditions. When compared to manual segmentation by experts, the results show good agreement. 

Bewley \etal \cite{bewley2012automated}  presented a technique for the automatic detection of kelps using AUV gathered images. The proposed method used local image features which are fed to Support Vector Machines (SVM) \cite{cortes1995support} to identify whether kelp is present in the image under examination. Comparison of several descriptors such as Local Binary Patterns (LBP) and Principal Component Analysis was carried out across multiple scales. This algorithm was tested on benthic data (collected from Tasmania in 2008), which contained 1258 images with 62,900 labels and 19 classes. The f1-score, which is the harmonic mean of precision and recall was used to evaluate the performance of their proposed method: 
\[f1 = 2\times \frac{precision \times recall}{precision + recall}\]
A maximum f1-score of 0.69 was reported for kelps. It was also suggested that practical systems can be built to assist scientists with automatic identification of kelps. They also concluded that results could be improved by using combinations at multiple scales, finding superior descriptors and by using more supplementary AUV data.  The study concluded that for a local geographical region, and for a particular species, sufficient generalization is possible.

This work was extended  in \cite{bewley2015hierarchical} for a multi-class classification problem in the presence of a taxonomical hierarchy. A local classifier was trained for each node of the hierarchy tree for LBP features and the classification results were compared through multiple hierarchy training methods. This algorithm achieved an f1-score of 0.75 for kelps and an overall mean f1-score of 0.197 for all 19 classes present in the dataset.

\subsection{Deep Learning for benthic marine Species Recognition}

In recent years, deep networks and off-the-shelf CNN features have become the first choice to tackle computer vision tasks. Only a handful of studies have developed benthic marine species recognition methods based on deep learning. Beijbom \etal \cite{beijbom2016improving} trained three and five-channel deep CNNs based on the CIFAR10 LeNet architecture \cite{krizhevsky2012imagenet} to improve the classification performance for coral and non-coral species. Reflectance and fluorescence images were registered together to obtain a five-channel image, which improved the classification performance by a significant margin. This was the first reported study to employ training of deep networks (from scratch) for benthic marine species recognition. 

Off-the-shelf CNN features \cite{razavian2014cnn} along with multi-scale pooling were first used for coral classification in \cite{mahmood2016coral} on the Moorea Labelled Coral (MLC) dataset, which is a challenging dataset introduced in \cite{beijbom2012automated}. This paper also explored a hybrid feature approach, combining CNN features with texton maps to further improve the classification accuracy on this dataset.  Class imbalance is an additional problem which refers to the disproportionate difference in the amount of points allocated to some classes compared to others. This is a common issue in benthic marine datasets, as some species are significantly more abundant than others.  To address the class imbalance, a cost-sensitive learning approach was studied in \cite{khan2017cost} using off-the-shelf CNN features for MLC dataset. In another study, features extracted from pre-trained deep networks were used to generate coral population maps for the Abrolhos Islands in Western Australia \cite{mahmood2016automatic}. This study reported a  trend of decreasing live coral cover in this region. This is consistent with the manual analysis of AUV images conducted by marine researchers \cite{smale2012regional,bridge2014variable}.

Deep residual networks (ResNets) are a special class of CNNs and are deeper, faster to train and easier to optimize than  previous CNN architectures \cite{he2015deep}. ResNets employ techniques such as residual learning and identity mapping for shortcut connections \cite{he2016identity}, which enables them  to overcome the limitations of traditional CNNs and outperform them in training speed and accuracy. ResFeats, features extracted from the output of convolutional layers of a 50-layer ResNet (ResNet-50), were reported to improve the performance of different image classification tasks in \cite{mahmood2016resfeats}, including coral classification on the MLC dataset. Although these features are computationally expensive large arrays, we chose to use the image representations extracted from the layers closer to the output end of  ResNet-50 to reduce computation cost and alleviate the  need for dimensionality reduction. 

\begin{table*}[]
\centering

\scalebox{0.9}{
\begin{tabular}{@{}p{4.5cm} p{5cm} cc@{}}
\midrule
Authors              & Methods                                                                                & Classes & Main Species                \\ \midrule
Marcos \etal \cite{marcos2005classification}         & Color histograms, local binary pattern (LBP)  and a 3-layer neural network                                       & 3       & Corals                 \\
Stokes and Deane \cite{stokes2009automated}      & Color histograms, discrete cosine transform and probability density based classifier                            & 18      & Corals, Macroalgae     \\
Pizarro \etal \cite{pizarro2008towards}         & Color histograms, Gabor filter response, scale-invariant feature transform (SIFT) and a voting based classifier & 8       & Corals, Macroalgae     \\

Beijbom \etal \cite{beijbom2012automated}             & Maximum response filter bank with SVM classifier                                                                & 9       & Corals, Macroalgae     \\
Denuelle and Dunbabin \cite{denuelle2010kelp}* & Haralick texture features with  Mahalanobis distance classifier                                                  & 2       & Kelp                   \\
Bewley \etal \cite{bewley2012automated}*        & Principal Component Analysis (PCA) and LBP descriptors   with SVM classifier                                      & 19      & Corals, Algae and Kelp \\
Bewley \etal \cite{bewley2015hierarchical}*          & Hierarchical classification with PCA  and LBP features                                                           & 19      & Corals, Algae and Kelp \\
Beijbom \etal \cite{beijbom2016improving}\(^{\bullet } \) & Deep neural network with reflectance   and fluorescence images       & 10      & Corals, Macrolagae     \\
Mahmood \etal \cite{mahmood2016coral}\(^{\bullet } \)       & Hybrid ( CNN + handcrafted) features  with a multilayer perceptron (MLP) network                                 & 9       & Corals, Macrolagae     \\
Mahmood \etal \cite{mahmood2016automatic}\(^{\bullet } \)     & Off-the-shelf CNN features  with SVM classifier                                                                  & 2       & Corals, Macroalgae     \\  \midrule
\end{tabular}}
\caption{A brief summary of methods for benthic image classification. \textit{ Key: *  have reported results on kelps and  \(^{\bullet } \) have used methods based on deep learning.} } 
\label{table:methods}
\end{table*}

\section{Methods and Results}
\label{S:3}

In this section, we outline the key components of our proposed method (Figure \ref{fig:outline}) and present the adopted experimental protocols.

\subsection{Datasets}

\subsubsection{Benthoz15 Dataset}

This Australian benthic data set (Benthoz15) \cite{bewley2015australian} consists of an expert-annotated set of geo-referenced benthic images and associated sensor data. These images were captured by AUV Sirius during Australia’s integrated marine observation system (IMOS) benthic monitoring program at multiple temperate locations  (Table \ref{table:benthoz}) around Australia \cite{williams2012monitoring}. Marine experts manually annotated each of these images according to the Collaborative and Automation Tools for Analysis of Marine Imagery and Video (CATAMI) classification scheme. For each image, up to 50 randomly selected pixels were hand labelled using the Coral Point Count with Excel Extensions (CPCe) software package \cite{kohler2006coral}. For each labelled pixel (point), a square patch of $224 \times 224$, centered at the labelled pixel is extracted. This patch is then used as an input for feature extraction.  These pixels were randomly selected using CPCe for manual annotations. A number of these pixels can be found on class boundaries, making the classification problem more challenging. The whole dataset contains 407,968 expert labelled points, taken from 9,874 distinct images collected at different depths and sites over the past few years.  There are 145 distinct class labels in this dataset, with pixel labels ranging from 2 to 98,380 per class. 33 out of these 145 classes belong to macroalgae (MA) species. 63,722 labelled points out of the total belong to the kelp class. Further details on the labeling methodology can be found in \cite{bewley2015australian}.

\begin{table*}
\centering

\begin{tabular}{@{}llll@{}}
\toprule
Site      & Survey Year      & \# of Pixel Labels & \# of Images \\ \midrule
Abrolhos Islands      & 2011, 2012, 2013 & 119,273   & 2,377     \\
Tasmania              & 2008, 2009       & 88,900    & 1,778     \\
Rottnest Island       & 2011             & 63,600    & 1,272     \\
Jurien Bay            & 2011             & 55,050    & 1,101     \\
Solitary Islands      & 2012             & 30,700    & 1,228     \\
Batemans Bay          & 2010, 2012       & 24,825    & 993       \\
Port Stevens          & 2010, 2012       & 15,600    & 624       \\
South East Queensland & 2010             & 10,020    & 501       \\ 
\midrule
Total & -            & 407,968   & 9,874      \\
\bottomrule
\end{tabular}
\caption{Benthoz15 data.}
 \label{table:benthoz}
\end{table*}

\subsubsection{Rottnest Island Dataset}

The Rottnest Island dataset was also collected by AUV Sirius and contains 297,800 expert labelled points, taken from 5,956 distinct images collected at different depths from five sites around Rottnest Island from 2010 to 2013 (Table \ref{table:rotto}).  Three out of the five sites are labelled north (15m, 25m and 40m depth) and  two as south (15m and 25m depth).  There are 78 distinct class labels in this dataset, with pixel labels ranging from 2 to 155,776 per class (Table \ref{class}). This makes the classification quite challenging. 25 out of these 78 classes belong to macroalgae species. 156,000 labelled points out of the total belongs to the kelp class.

\begin{table}
\begin{center}
\scalebox{1}{
\begin{tabular}{ @{} cccc@{}  }
\toprule
  Survey Year & \# of Images  & \# of Pixel Labels   & \# of Classes\\
\midrule
 2010 &	1,680  &	84,000 & 61\\ 
 2011 &	1,680 &	84,000& 55\\ 
 2012 & 1,033 &51,650 & 44\\ 
 2013& 	1,563 &78,150 & 55\\ 
 \midrule
 Total& 5,956&297,800 & 78\\ 
\bottomrule
\end{tabular}
}
\caption{Rottnest Island data. } 
\label{table:rotto}

\end{center}
\end{table}

\begin{figure*}
\begin{center}
   \includegraphics[width=1\linewidth,height=0.6\linewidth,keepaspectratio]{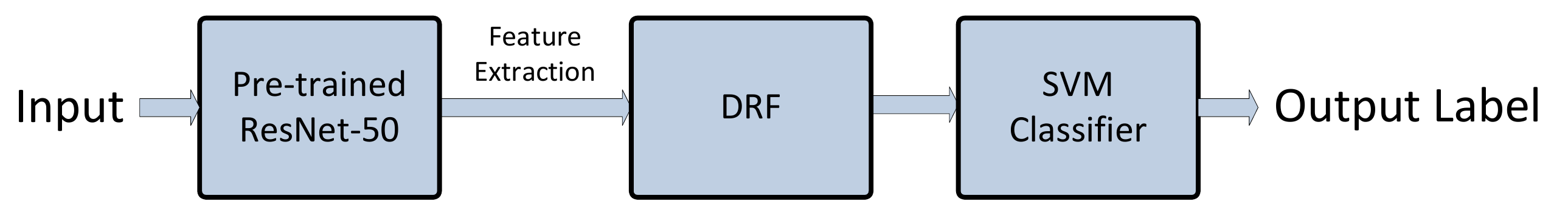}
\end{center}
   \caption{  The block diagram of our proposed framework. }
\label{fig:outline}
\end{figure*}

\subsection{Classification Methods}

\begin{figure*}
\begin{center}
   \includegraphics[width=1\linewidth,height=1.5\linewidth, keepaspectratio]{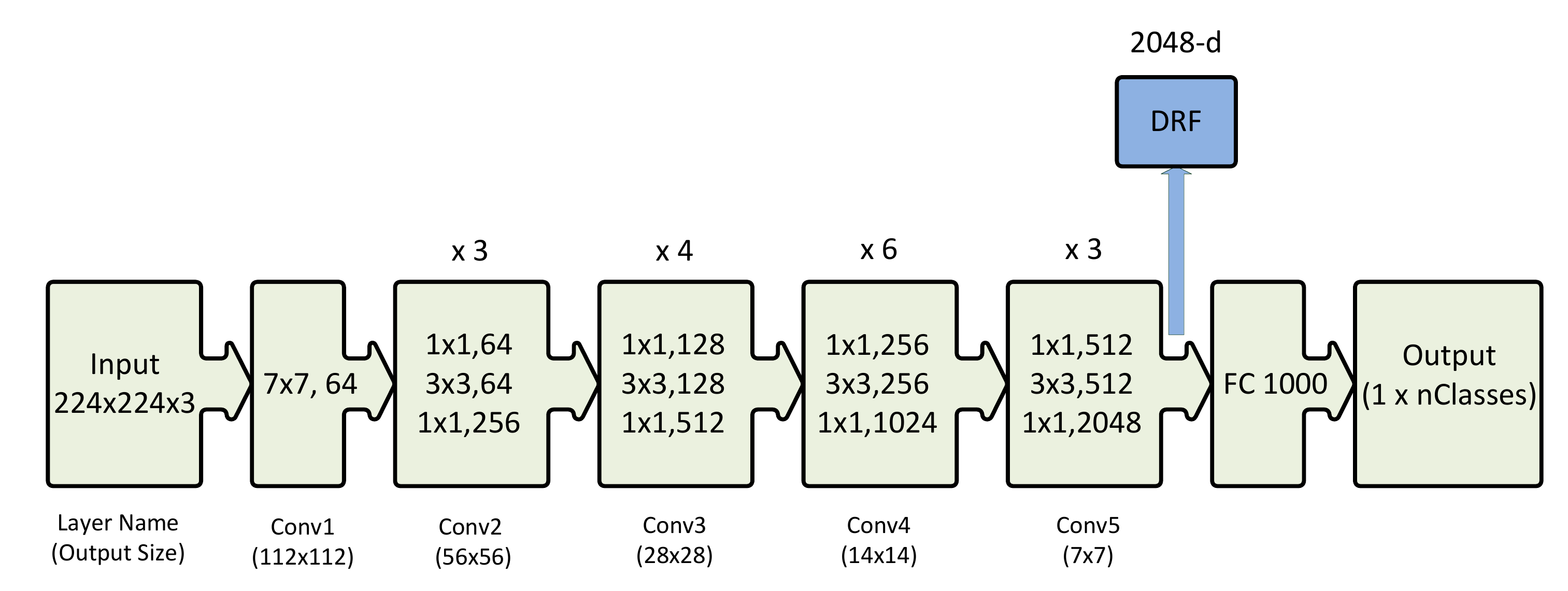}
\end{center}
   \caption{ResNet-50 architecture \cite{he2015deep} shown with the residual units, the size of the filters and the outputs of each convolutional layer. DRF extracted from the last convolutional layer of this network is also shown. \textit{Key: The notation \(k\times k, n\) in the convolutional layer block denotes a filter of size k and  n channels. FC 1000 denotes the fully connected layer with 1000 neurons. The number on the top of the convolutional layer block represents the repetition of each unit. nClasses represents the number of output classes.  }}

\label{fig:res50}
\end{figure*}

Deep residual features are extracted from the output of the last convolutional block of a 50-layer deep residual network (ResNet-50) \cite{he2015deep} that is pre-trained on ImageNet.  Figure \ref{fig:res50} shows the architecture of the ResNet-50 deep network which we have used for feature extraction. The ResNet-50 is made up of five convolutional blocks stacked on top of each other (Figure \ref{fig:res50}). The convolutional blocks of a ResNet are different from those of the traditional CNNs because of the introduction of a shortcut connection between the input and output of each block.  Identity mappings when used as shortcut connections in ResNets  \cite{he2016identity}, can lead to better optimization and reduced complexity.  This in turn allows one to use deeper ResNets which are faster to train and are computationally less expensive than the conventional CNNs \ie VGGnet \cite{simonyan2014very}. 

The image representations extracted from the  fully connected layers of deep networks pre-trained on ImageNet \cite{razavian2014cnn}  capture the overall shape of the object contained in the region of interest. The features extracted from the deeper layers encode class specific properties (\ie shape, texture and color) and give superior classification performance as compared to features from shallower layers \cite{zeiler2014visualizing}. Hence, we propose to extract the features from the output of the last convolutional block of ResNet-50 (Figure \ref{fig:res50}). The output of the Conv5 block is a $7\times7\times2048$ dimensional array and is used as input of the FC-1000 layer. This large array is however, first converted to a 2048-dimensional vector by using a max-pool layer. We extract this 2048-dimensional vector and name it DRF. We do not use the FC-1000 layer for feature extraction because it is used as an output layer to classify the 1000 classes of the ImageNet dataset, which was used to pre-train this network. Our feature extraction method is different from the conventional method employed in previous deep networks such as VGGnet. The presence of multiple fully connected layers in the VGGnet makes the feature extraction straightforward. The only fully connected layer in ResNet is class specific to the ImageNet dataset. Therefore, we proposed to use the output of the last convolution block for DRF extraction.

There are three different approaches described in \cite{silla2011survey} to deal with the hierarchical classification problem: 

\begin{enumerate}
\item \textbf{Flat Classification:} This approach ignores the hierarchy and treats the problem as a parallel multi-class classification problem.
\item \textbf{Local Binary Classification:} A binary classifier is trained for every node in the hierarchical tree of the given problem. 
\item \textbf{Global Classification:} A single classifier is trained for all classes and the hierarchical information is encoded in the data. 
\end{enumerate}

\begin{figure}
\begin{center}
\scalebox{1}{
   \includegraphics[width=1\linewidth,height=1.5\linewidth, keepaspectratio]{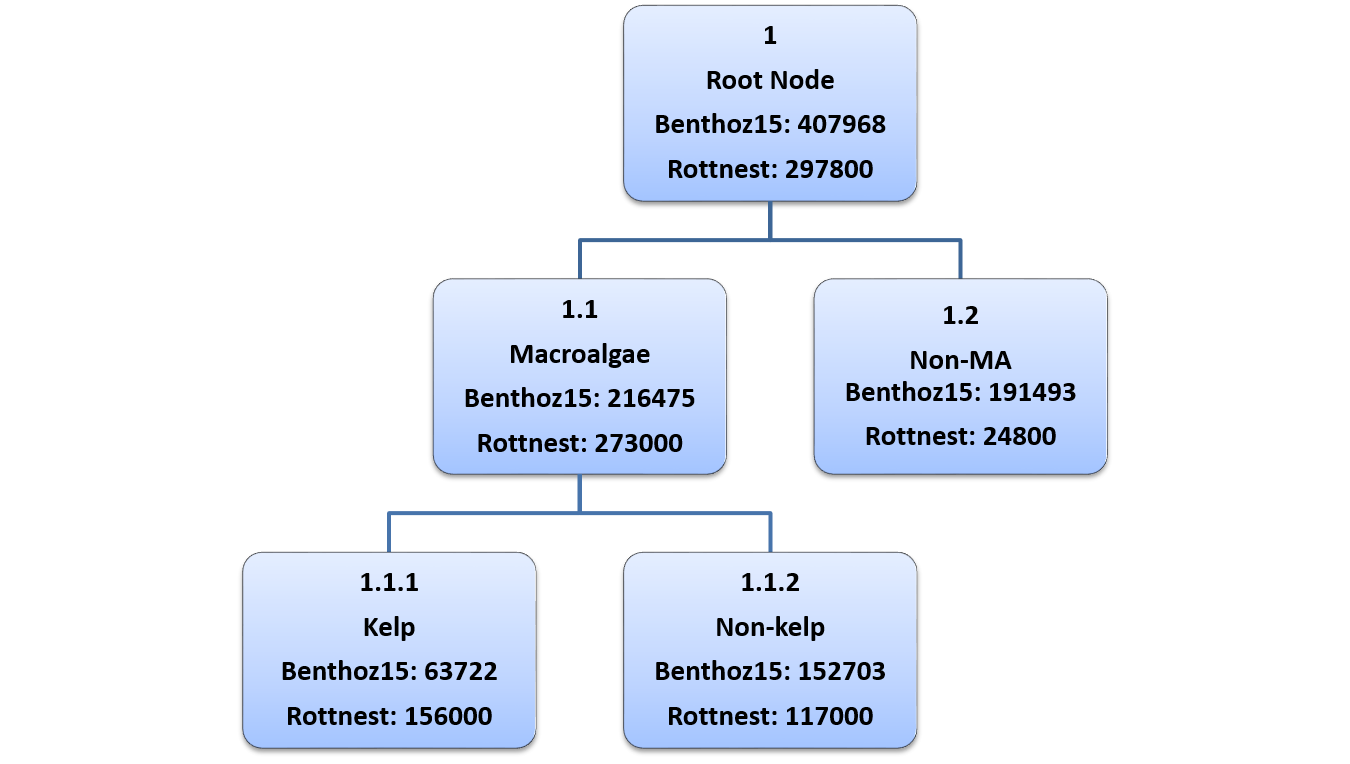}}
\end{center}
   \caption{\small{Hierarchy tree for kelps in our benthic data. In each node, the first line shows the node number, 2$^{nd}$ line shows the name of the specie, and 3$^{rd}$ and 4$^{th}$ lines show the number of labels belonging to that particular species in Benthoz15 and Rottnest Island data respectively.}}

\label{fig:tree}
\end{figure}

We have used the local binary classification technique in this paper to identify kelps from other taxa. This approach is easier to implement and more useful when all the nodes in the hierarchy are not labeled to a specific leaf node level. For example, some macroalgae are not labeled to the species level in the Benthoz15 dataset \cite{bewley2015australian}. Moreover, this approach also allows for the use of different features, training sets and classifiers for each node of the hierarchy tree. The hierarchy tree for kelps is shown in Figure \ref{fig:tree}. 

\subsection{Training and Testing Protocols}

In this paper, two training approaches are used, namely \textit{inclusive training} and \textit{sibling training}. In the inclusive training method, all the non-kelp samples from the entire dataset are treated as negative samples \ie nodes 1.2 and 1.1.2 in Figure \ref{fig:tree}. However in the sibling training method, only those non-kelp samples are considered as negative which comes under the macroalgae node \ie node 1.1.2 in Figure \ref{fig:tree}. We use a linear Support Vector Machines (SVM) \cite{cortes1995support} classifier because it has shown excellent performance with  features extracted from deep networks \cite{razavian2014cnn}. We use the SVM classifier in a one-vs-all configuration with a linear kernel. We perform 3-fold cross validation  within the training set to optimize the SVM parameters and mean performances are reported in Section \ref{S:3}. 

\subsection{Image Enhancement and Implementation Details}

We applied color channel stretch on each image in the dataset to reduce the effect of underwater color distortion phenomenon. We calculated the  averages of the lowest 1\% and the highest 99\% of the intensities for each color channel. The average of the lowest 1\% intensities  was subtracted from all the intensities in each respective channel and the negative values were set to zero. These intensities were then divided by the average of the highest 99\% of the intensities. This process enhanced the color information of benthic marine images.

For feature extraction, we used a pre-trained ResNet-50 \cite{he2015deep} deep network architecture in our experiments. We used the publicly available model of this network, which was pre-trained on the ImageNet dataset. We implemented our proposed method using MatConvNet \cite{vedaldi2015matconvnet} and the SVM classifier using LIBLINEAR \cite{fan2008liblinear} (Figure \ref{fig:outline}).

\subsection{Experimental Settings and Evaluation Criteria}

70\% of images from each geographical location were used to form the training set for experiments carried out on the Benthoz15 dataset. However, for Rottnest Island data, the images from years 2010, 2011 and 2012 are included in the training set and the images from year 2013 form the testing set. We   performed our experiments with three different   classification approaches: flat classification and local binary classification with both inclusive and sibling training policies.   
The overall classification accuracy is not an effective measure of binary classifier performance for datasets  exhibiting a skewed class distribution. Therefore, to evaluate the performance of our classifier, we have used four evaluation criteria: overall classification accuracy, mean f1-score (the average of f1-scores of each class involved in the test data), precision and recall values of kelp. 

\subsection{Classification Results}

In this section, we report the results of three different types of features for the three training methods on the two datasets: \textbf{(i)} Maximum Response (MR) filter and texton maps of \cite{beijbom2012automated} as baseline handcrafted features. We used a publicly available implementation of this method; \textbf{(ii)} CNN features extracted from a VGG16 network pretrained on ImageNet dataset \cite{simonyan2014very}; \textbf{(iii)} Our proposed DRFs extracted from a pretrained ResNet-50. 

Classification by the DRF method always outperformed the traditional CNN features and MR features in both datasets as it consistently showed higher accuracy, higher f1 scores, higher precision of kelps and higher kelp recall than previously used features. Additionally,  hierarchical classification (sibling and inclusive) in comparison to flat classification, also improved f1-score and recall of kelps while providing lower training times. The sibling training method achieved the highest f1-score for both  datasets. Because f1-score is an evaluation metric based on both precision and recall, we recommend the sibling training method as the top performing practical method for classification and automated coverage analysis of kelps. 

\subsubsection{Benthoz15 Dataset}

To highlight the superior classification performance of DRF, we have included a comparative study among DRF and the traditionally used CNN features extracted from VGGnet \cite{simonyan2014very} and MR features (Table \ref{table:res1}).  The DRF method performs better than both the features for all three classification experiments. The lowest overall accuracy was achieved by the flat multi-class classification method (57.6\%). Additionally, a very low mean f1-score of 0.05 was observed, since many classes among the total 145 had very few samples for training and testing. Nonetheless, the flat classification method achieved the highest precision (71\%) for kelps among all the three methods. Out of every 100 kelp samples, this method correctly identifies 71 samples as kelps. However, this method resulted in the worst recall value of 65\% (Table \ref{table:res1}). 

The best classification accuracy is achieved with the inclusive training method (90\%) for which all the non-kelp samples are bundled together in the negative class. This training scheme achieves a mean f1-score of 0.79 which is similar to the highest f1-score of 0.80 obtained using the sibling training method (Table \ref{table:res1}). 

The sibling training method is  more challenging  as compared to the inclusive training method because the negative samples only include macroalgae classes and some of these classes are very similar to kelp in appearance. This accounts for a drop in classification accuracy from 90\% to 83.4\%. However the sibling training method resulted in the highest mean f1-score (0.80) and recall value (78\%) for kelp.  Moreover, statistical testing supports the hypothesis that all three DRF classifiers are   better than their VGG and MR counterparts at  significance level of 0.05. For each DRF feature $X$  and competing feature $Y \in (MR, VGG)$, we did a paired t-test over randomly chosen image samples ($N=50,000$), using the SVM classifier. Statistical results showed that, for each pairing of features $(X,Y)$, feature X gave better classification than feature Y at the 0.05 significance level. The calculated p-value was less than 0.05 which rejected our null hypothesis that both classifiers show similar performance.

\begin{table*}[]
\centering
\rotatebox{0}{

\scalebox{1}{
\begin{tabular}{@{}ccccc@{}}
\toprule
 Method & Accuracy (\%) & Mean f1-score & Precision of Kelps (\%) & Recall of Kelps (\%)\\ \midrule
 MR: Flat  &   51.6$\pm 0.3$    &     0.03$\pm 0.00$   &        64$\pm 0.5$       &      59$\pm 0.5$   \\
 MR: Inclusive       &    82.8$\pm 0.4$   &     0.70$\pm 0.03$   &        43$\pm 0.0$       &      69$\pm 0.0$     \\
 MR: Sibling         &    79.6$\pm 0.3$  &      0.72$\pm 0.02$   &      55$\pm 0.0$      &     73$\pm 0.0$        \\
\midrule 

VGG: Flat  &   54.4$\pm 0.6$    &     0.03$\pm 0.01$   &        67$\pm 0.5$       &      63$\pm 0.5$   \\
VGG: Inclusive       &    89.0$\pm 0.5$   &     0.75$\pm 0.02$   &        47$\pm 0.0$       &      73$\pm 0.0$     \\
VGG: Sibling         &    82.1$\pm 0.4$  &      0.76$\pm 0.01$   &      57$\pm 0.0$      &     75$\pm 0.0$        \\
\midrule
DRF: Flat         &   57.6$\pm 0.5$   &     0.05$\pm 0.02$     &      \textbf{71$\pm 1.0$  }      &     65$\pm 1.0$     \\
DRF: Inclusive       &    \textbf{90.0$\pm 0.07$ }      &       0.79$\pm 0.02$         &          58$\pm 0.0$          &       73$\pm 0.0$           \\
DRF: Sibling         &     83.4$\pm 0.2$      &         \textbf{0.80$\pm 0.01$ }    &    65$\pm 0.0$         &        \textbf{78$\pm 0.0$ }       \\ \bottomrule
\end{tabular}}}
\caption{A comparison of flat, inclusive and sibling classification methods for kelp classification on Benthoz15 dataset for MR, VGG and DRF methods.  The flat classification focuses on all the classes present in the dataset whereas the inclusive and sibling classification only includes kelps and non-kelps. Mean f1-score corresponds to the average of the individual f1-score of each class involved in the experiment. Best scores are shown in bold font.}
\label{table:res1}
\end{table*}

\subsubsection{ Rottnest Island Dataset}

The DRF was then applied to the Rottnest Island data and once again confirmed that the DRF outperformed the VGG and MR features for all the classification experiments (Table \ref{table:res2}).  The hierarchical methods performed better than the flat classification method for all  evaluation criteria except for precision. However, the recall value achieved by this method is the worst. This is consistent with the results obtained on Benthoz15 dataset. The mean f1-score for flat classifier (0.03) is again very low given the fact that all 78 classes are classified at the same time. The sibling training method comes out as the best method with respect to accuracy (77.2\%), mean f1-score (0.76) and recall value (79\%) of kelps. Moreover, the sibling training method is also the fastest method because it has less negative examples than the inclusive method.

Fine-tuning a deep network is also a popular approach for transfer learning \cite{azizpour2015generic}. We also  compared our proposed method with fine-tuning. Fine-tuning a ResNet-50 on Rottnest Island data achieved an overall classification accuracy of 58.8\% as compared to the 59.0\% achieved by our proposed method.  For Benthoz15 dataset, fine-tuning a ResNet-50 resulted in an overall classification accuracy of 57.1\% which is 0.5\% lower than our proposed method. The performance change was marginal for both datasets. Hence, we concluded that the classification accuracy achieved by both methods on benthic marine datasets is comparable. One important aspect to compare is the computational time required by these two approaches. The time needed to extract off-the-shelf features from a ResNet and classify them using an SVM classifier is far less than the time required to fine-tune a 50 layer ResNet on a dataset as large as 297,800 input images. Our proposed method requires a few hours to run. However, fine-tuning a ResNet-50 with Rottnest Island dataset takes at least 2 days on an Nvidia Titan-X GPU. Given these considerations, we selected our proposed method over fine-tuning a ResNet with a marine dataset approach.

\begin{table*}[]
\centering
\rotatebox{0}{

\scalebox{1}{
\begin{tabular}{@{}ccccc@{}}
\toprule
Method & Accuracy (\%) & Mean f1-score & Precision of Kelps (\%) & Recall of Kelps (\%)\\ \midrule
 MR: Flat  &   52.9$\pm 0.4$      &     0.02$\pm 0.00$       &        90$\pm 2.0$          &      62$\pm 1.0$       \\
 MR: Inclusive    &    73.2$\pm 0.6$      &     0.70$\pm 0.01$   &        77$\pm 0.0$       &       74$\pm 0.0$         \\
 MR: Sibling    &  71.7$\pm 0.4$     &       0.71$\pm 0.01$      &      80$\pm 0.0$           &       73$\pm 0.0$         \\
\midrule
VGG: Flat  &   58.6$\pm 0.6$      &     0.02$\pm 0.01$       &        95$\pm 1.5$          &      65$\pm 1.0$       \\
VGG: Inclusive    &    74.7$\pm 0.4$      &     0.74$\pm 0.02$   &        81$\pm 0.0$       &       75$\pm 0.0$         \\
VGG: Sibling    &  74.5$\pm 0.3$     &       0.73$\pm 0.02$      &      84$\pm 0.0$           &       75$\pm 0.0$         \\
\midrule
DRF: Flat         &   59.0$\pm 0.7$   &      0.03$\pm 0.01$    &    \textbf{95$\pm 1.0$ }         &     66$\pm 1.0$     \\
DRF: Inclusive       &    75.0$\pm 0.5$       &       0.75$\pm 0.01$         &          82$\pm 0.0$           &       75$\pm 0.0$           \\
DRF: Sibling         &     \textbf{77.2$\pm 0.4$  }    &  \textbf{0.76$\pm 0.02$ }      & 86$\pm 0.0$   &        \textbf{79$\pm 0.0$ }    \\
\bottomrule
\end{tabular}}}
\caption{ A comparison of flat, inclusive and sibling classification methods for kelp classification on Rottnest Island dataset for MR, VGG and DRF methods.  The flat classification focuses on all the classes present in the dataset whereas the inclusive and sibling classification only includes kelps and non-kelps. Mean f1-score corresponds to the average of the individual f1-score of each class involved in the experiment. Best scores are shown in bold font. }
\label{table:res2}
\end{table*}

One of many challenges in benthic cover estimations through image analysis is the large amount of time required to manually classify the imagery. The average time for manual annotation with 50 sample points per image is 8 minutes. A trained marine expert can annotate up to 8 images per hour. The proposed method is significantly less time consuming as it results in an annotation rate of 1800 images per hour using a Nvidia Titan-X GPU. This is approximately 225 times faster than manual annotation by experts. Nonetheless, note that the proposed machine learning algorithm is only classifying ‘kelp’ vs ‘non kelp’. Although it is faster, it is not yet trained to classify 145 potential benthic classes. This paper evaluates the technique for a single class and presents a way forward to develop the methodology for other classes and faster processing times, which will allow scientists to promptly analyze changes in benthic community composition.

\subsection{Kelp Coverage Analysis}

\begin{figure*}
\begin{center}
\scalebox{1}{
   \includegraphics[width=1\linewidth,height=1.5\linewidth, keepaspectratio]{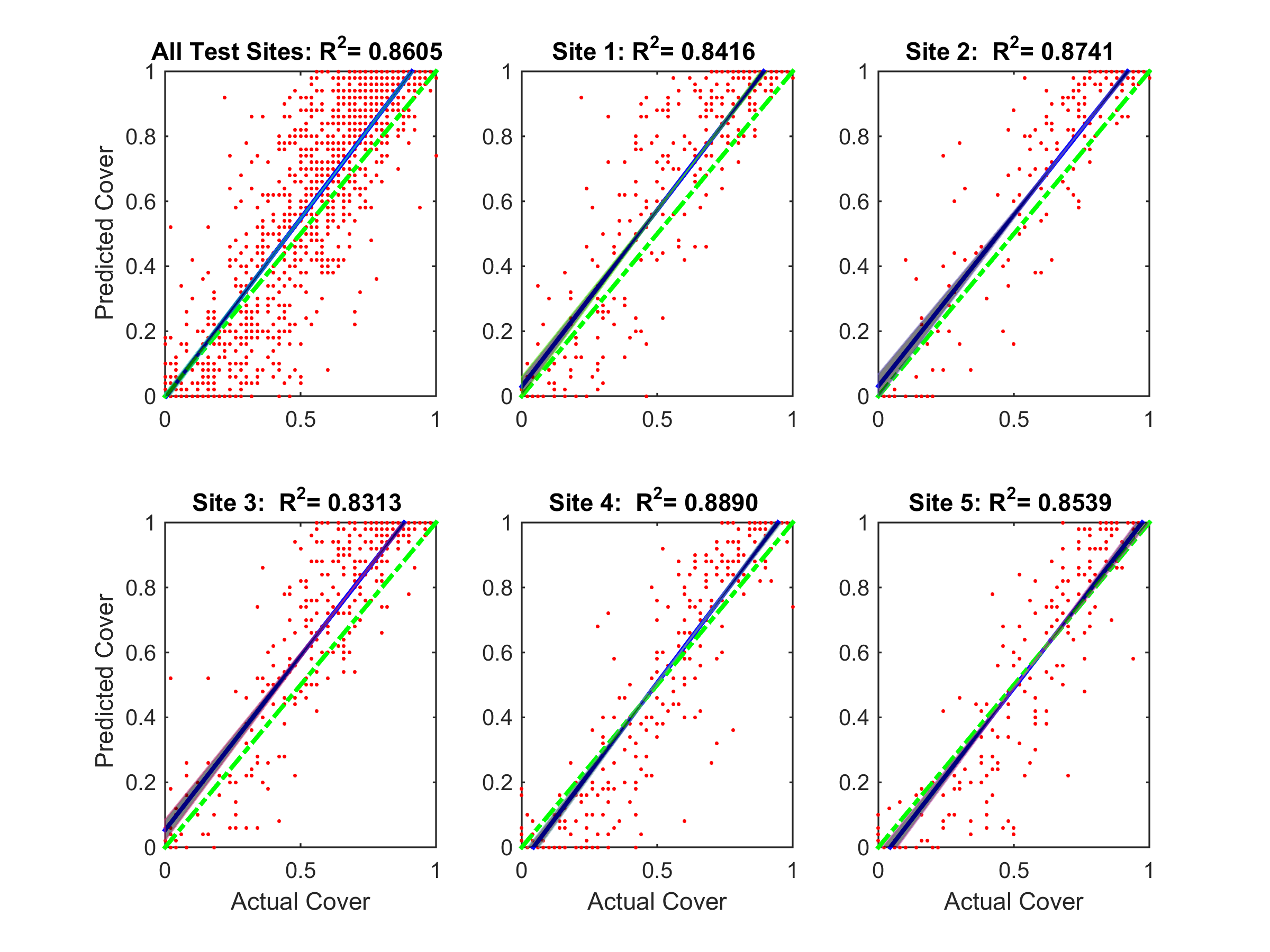}}
\end{center}
   \caption{Coverage estimation scatter plots for Rottnest Island Data for the DRF: Sibling Training experiment. Each dot indicates the estimated cover and the actual cover per image. The dashed green line represents the perfect estimation. The blue line on each plot is the linear regression model and the shaded area represent the 95\% confidence intervals. The first plot is the aggregated plot of the remaining plots of the five sites included in the 2013 test data. $R^2$ value for each sub-plot is shown in the respective title.}
\label{fig:kelp}
\end{figure*}

We  extended our method to estimate kelp cover for the Rottnest Island dataset. The expert identified coverage was calculated by aggregating the pixel level ground truth labels in every image. We calculated the estimated kelp coverage by aggregating the predicted labels for the same locations for which the expert labels were available. Kelp cover estimated by the annotations generated by our proposed method was compared to the cover based on expert classification (Figure \ref{fig:kelp}; Table \ref{table:allcover}). Scatter plots were generated for each of five sites and all the data included in the 2013 test set.
An important application of our proposed method is to estimate the population trends of kelp across spatial and time scales. To accomplish this task, we split the Rottnest Island data into sites and trained a classifier on this basis instead of years. The three sites from the north constitute the training set and the two southern sites form the test set.

The first sub-plot in Figure \ref{fig:kelp} shows  kelp coverage for all of the data included in the test set. The slope of the line generated by linear regression is very close to the ideal case. This highlights the robustness of our proposed algorithm. The remaining sub-plots show  kelp coverage for each of the five sites. These sub-plots show a good agreement between the annotations generated by our proposed method and the annotations provided by the human experts (Table \ref{table:allcover}). Moreover, we also calculated the R-squared ($R^2$) value for each plot to show correlation between the actual and predicted cover. Our proposed method achieved a high $R^2$ value for each individual site and then all sites combined. It is important to note that the DRF classification seems to over-fit kelp cover at high percentages of cover and to under-fit kelp cover at lower ones. 

\begin{figure}
\begin{center}
\scalebox{1}{
   \includegraphics[width=1\linewidth,height=1.5\linewidth, keepaspectratio]{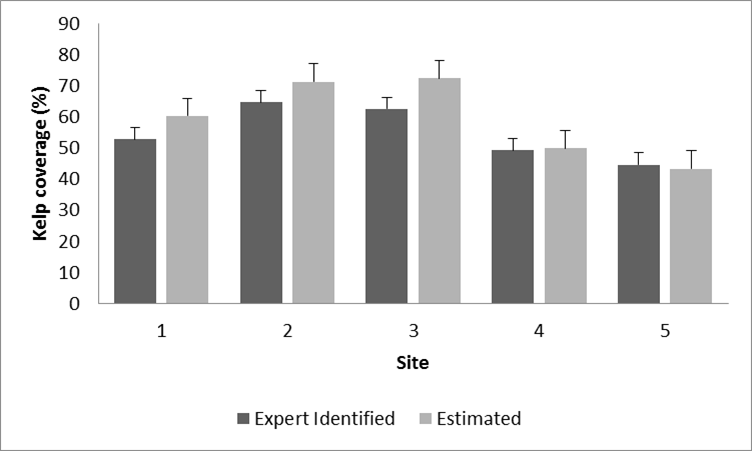}}
\end{center}
   \caption{Expert identified and estimated kelp coverage for all five sites of Rottnest Island data for year 2013. }

\label{fig:kelp_2013}
\end{figure}

 The estimated kelp coverage is not significantly different from the coverage calculated by the experts from the ground truth labels (Figure \ref{fig:kelp_2013}). This indicates the robustness of  our proposed method for estimating kelp coverage.  These results are beneficial to marine scientists since many surveys focus on estimating kelp coverage, which is an important metric to indicate the health of kelp forests.

\begin{figure*}
\begin{center}
\scalebox{1}{
   \includegraphics[width=1\linewidth,height=1.5\linewidth, keepaspectratio]{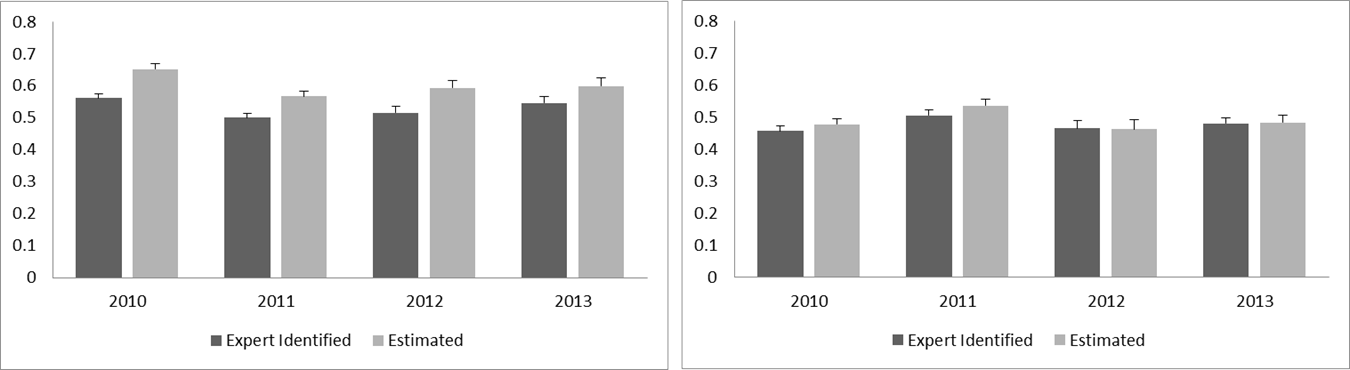}}
\end{center}
   \caption{Expert identified and estimated kelp coverage for the two southern sites of the Rottnest Island data. \textit{Left: Site 2, Right: Site 4}.}

\label{fig:kelp_pop}
\end{figure*}

Figure \ref{fig:kelp_pop} shows the expert identified and estimated percent cover of kelp across years of sites 2 and 4. For site 2, a slight over estimation of kelp cover by the DRF classification is visible, however no distinct trend of change across years is observable in either manual or automatic classification. On the other hand, the estimation of kelp cover for site 4 shows no overestimation and similarly to site 2, no trend change in kelp cover over the years.

\begin{table}[]
\centering

\scalebox{0.7}{
\begin{tabular}{@{}ccccc@{}}
\toprule
Site &Depth and Location& Expert Identified (\%) & Estimated (\%) &  $R^2$    \\ \midrule
1 &  15m North & 52.65             & 60.19     & 0.84 \\
2  &  15m South  & 64.64             & 71.23     & 0.87 \\
3   &  25m North & 62.44             & 72.32     & 0.83 \\
4  &  25m South  & 49.24             & 49.78     & 0.89 \\
5  &  40m North  & 44.60             & 43.28     & 0.85 \\ \bottomrule
\end{tabular}}
\caption{Expert identified and estimated kelp coverage for all five sites of Rottnest Island data for year 2013 along with the $R^2$  values.}
\label{table:allcover}
\end{table}

\section{Discussion}
\label{S:4}

The use of AUVs to survey  benthic marine habitats has allowed scientists to investigate remote locations such as off-shore and deep sites, which are beyond the limits of traditional SCUBA diving. Nonetheless, the efficiency of image collection does not match the availability of data for ecological analysis, as image classification  is time consuming and costly given that it is performed manually by marine experts. Additionally, manual classification has other disadvantages such as observer discrepancies and biases. Automated analysis of imagery is thus  essential to fully benefit from the advantages of remote surveying technologies such as AUV’s. In this study, we have addressed this problem by evaluating a machine learning automated image classification method using Deep Residual Features (DRF) for a key marine benthic species: the kelp \textit{Ecklonia radiata}. 

We have demonstrated that the image representations extracted from pre-trained deep residual networks can be effectively used for benthic marine image classification in general and kelps in particular. These powerful and generic features outperform  traditional off-the-shelf CNN features, which have already shown superior performance over conventional hand-crafted features \cite{mahmood2016coral, razavian2014cnn}. The sibling and inclusive hierarchical training  methods further enhance  performance when compared to flat multi-class classification methods. The sibling and inclusive training methods show comparatively similar performance. However, the sibling method is superior because it has lower training time than the inclusive method. Furthermore, estimations of kelp cover by automated DRF classification  closely resemble those of manual expert classifications with the added advantage of faster processing times. This work provides evidence that automatic annotations may save resources and time while providing effective estimates of  benthic cover.

This method was also applied on a dataset to compare kelp coverage for multiple sites, across three depths and for a consecutive time series of four years (2010-2013) at Rottnest Island. The patterns observed showed differences in percent cover of the kelp \textit{Ecklonia radiata} between sites (with higher percentage cover of kelp in shallower sites compared to deeper sites) and no considerable change of kelp cover across years. These trends were similar to those observed by manually classified data once more confirming the usefulness of automated image classifying methods and the ability to use them for ongoing monitoring of kelp beds with AUV technology. 

In this study, we found no evidence of catastrophic loss of kelp over the years at any of the sites surveyed at Rottnest Island. These results are comparable to previous estimates of change in \textit{E. radiata} cover across depth in Australia, performed with manually classified images \cite{marzinelli2015large}. They are in contrast with trends of significant and continuous kelp decline reported in the region after an extreme marine heatwave which resulted in widespread mortality of benthic species including corals, seagrasses, invertebrates and kelp \cite{wernberg2016climate}. The loss of kelp in Western Australia resulted in a range contraction of 100 km \cite{wernberg2016climate} and in crab and scallop fishery closures of benthic species associated with kelp habitat. Importantly, the kelp loss was reported in habitats shallower than 15 m, with little attention to the response of deeper habitats to the heatwave \cite{smale2012regional}. This may be why our results contrast with studies reporting catastrophic loss of kelp, since our shallowest locations were at 15 m of depth, and most in situ studies take place even shallower (about 12 m). Additionally, all our sites were located off-shore (even the shallow ones), which may indicate that off-shore sites are less impacted by environmental pressures. This may be due to the lack of other environmental disturbances that coastal habitats are exposed to, due to their distance to shore and human populations. The interaction of several disturbances has been shown to cause ecological responses such as wide spread mortality of marine benthic species \cite{fraser2014extreme}. Kelps growing offshore and in deeper locations (> 15 m of depth) appear  to be less impacted by extreme warming in contrast to coastal shallow reefs \cite{anita2019}. As a result of the catastrophic consequences that extreme climatic events may have on key habitat building species, such as kelp, deeper marine regions have been identified as potential refugia for shallow marine species \cite{graham2007deep, lesser2009ecology, kahng2010community}. This emphasizes the importance of AUV surveys to provide information on offshore and deep locations which may be influenced by different factors to their inshore counterparts \cite{smale2012regional}. The use of automated image analysis for processing AUV images will streamline the processing of these images to efficiently identify patterns observed in deep and remote locations and compare them with patterns observed in shallow and inshore sites. 

The rapid characterization of ecological changes is crucial in light of the catastrophic threats to marine biodiversity posed by the rise of extreme climatic events driven by climate change and other anthropogenic stressors. Technology has enabled the rapid collection of images even in remote locations through autonomous underwater vehicles, remotely operated vehicles, automated cameras and even satellite imagery. The subsequent annotation of such imagery is typically time consuming and consequently, the automation of marine species classification from digital images has become a priority. This study focuses on the kelp species \textit{ E. radiata}, which is the dominant habitat builder of temperate reefs in Australia, though automated classification of marine species has been applied to other important marine species. For example, progress in automated tropical coral identification has resulted in accurate classification the level of genera \cite{beijbom2015towards} . Other successful automated classification techniques for coral reefs include the collection of multifaceted data, minimum manual classification effort (around 2\% of pixels) and machine learning techniques which result in cm-scale benthic habitat maps of high taxonomic resolution and accuracy of up to 97\% \cite{chennu2017diver}. Similarly, in pelagic species such as fish automated classification has advanced rapidly, with automated fish detection and identification algorithms also measuring basic fish morphological features such as total length \cite{williams2016automated,shortis2016progress}. In contrast, automated methods for identification of marine macroalgae from benthic images still result in low agreement \cite{beijbom2015towards}, highlighting the need for more research into unequivocal definitions of algal groups for image classification.

Although the proposed DRF classification method allowed us to compare kelp cover in different sites and across different years providing marginal differences with the estimations from manual annotations, there were some errors associated with the proposed technique. We observed an over-prediction of kelp at high percentage cover and under-prediction at low cover. Nonetheless, the over prediction was smaller when data was divided per site and in some sites was negligible (4 and 5). Overall, the estimated kelp cover closely resembles manual classification and taking into consideration the cost effectiveness of automated DRF classification methods, the benefits of the automated classification method out-weight the drawbacks. As such, automated classification of kelp from AUV-derivated images constitute a cost-effective method for estimations of kelp abundance across space and time.

\begin{figure*}
\begin{center}
\scalebox{0.6}{
   \includegraphics[width=1\linewidth,height=1.5\linewidth, keepaspectratio]{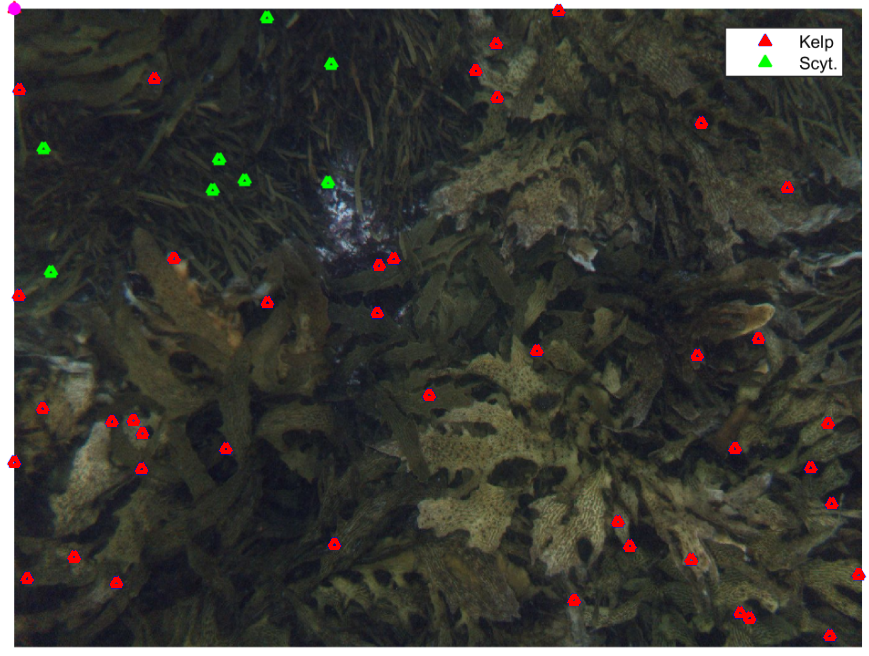}}
\end{center}
   \caption{An example image from Rottnest Island Dataset with manual annotations showing  similarity in appearance between \textit{Scytothalia dorycarpa} (green)  and  the kelp \textit{Ecklonia radiata} (blue).  }

\label{fig:comp}
\end{figure*}

A comparison of the best overall accuracies of hierarchical classification across the two used datasets shows that both the sibling and inclusive DRF classifiers has shown better classification accuracy on Benthoz15 dataset as compared with Rottnest Island dataset. For example, the inclusive DRF classifier for Benthoz15 dataset (Table \ref{table:res1}) has an absolute gain of 15\% over the respective classifier for the Rottnest dataset (Table \ref{table:res2}). This substantial difference is possibly due to the high presence of the brown algae \textit{Scytothalia dorycarpa} in the Rottnest Island data. \textit{Scytothalia dorycarpa} is very similar to kelp in appearance and usually occurs in areas of the sea floor with high cover of kelp. Therefore, marine scientists may mis-classify it as kelp in poor quality images. This misclassification is possible if the point falls on the edge of \textit{Scytothalia dorycarpa}, where the boundary between the two species is not clear.  The expert misclassification of \textit{Scytothalia dorycarpa} as kelp may also explain the over-prediction of kelp by the DRF classification method at high percentage cover. The over-prediction of the automated classification is actually an overestimation of the kelp cover by the manual annotation method. The subjectivity in the classification is removed by the automated analysis, which uses several features to classify kelp. Figure \ref{fig:comp} illustrates the similarity of appearance of these two species.

Poor quality images (low light and resolution) will also affect the manual classification of other classes of algae such as ‘turf matrix’, ‘fine branching red algae’ or other canopy forming brown algae. These and other algae classes are not as common as kelp at the sites surveyed at Rottnest Island. Thus, misclassification associated to manual annotations may also explain the over prediction of kelp at low percentage covers. At low  cover of kelp, a turf and foliose matrix of red algae occurs on the rocks. In areas of low kelp cover it is easy for an expert to distinguish kelp from other classes, but perhaps due to the imbalance of data for training the classifier sometimes other classes are classified as kelp resulting in over-prediction by the DRF classification method. These issues highlight the need for larger training datasets for deep learning based automatic annotation. Extensive and comprehensive training sets will allow for better classifier training  and give the opportunity to increase the amount of biota classified automatically (e.g. other algae species, corals, sponges, invertebrates such as sea urchins and lobsters). Future work will explore multi-class classification of benthic marine species across diverse benthic habitats so methods based on deep learning algorithms can be applied to numerous ecological problems that include other benthic marine species. Scientists who use data extracted from image classification should keep these considerations in mind when manually annotating images since these datasets are extremely valuable for deep learning based automatic classification.

\section{Conclusion}
\label{S:5}

The aim of this study was to investigate deep learning techniques for automatic annotation of kelp species in a complex underwater scenery. Towards this end, we evaluated a Deep Residual Features (DRF) based method to carry out this task and showed it outperformed the widely adopted off-the-shelf CNN based classification. We also established that hierarchical classification with the sibling method gave superior results compared to the flat multi-class approach with the added advantage of faster training times. Our results suggest that the proposed automatic kelp annotation method can significantly reduce the number of human-hours spent in manual annotations. Additionally, our proposed method can enhance the effectiveness of AUV monitoring campaigns by facilitating the early detection of changes in the population of key species though rapid image processing times, as demonstrated with examples from the Rottnest Island dataset. To conclude, the proposed DRF based automatic annotation of benthic images   is to this date the most accurate machine learning technique for estimation of kelp cover.

\vspace{6pt} 



\authorcontributions{Ammar conceived the idea, designed the experimental protocols and led the writing of the manuscript. Anita and Renae collected the data and provided manual annotations. Mohammed and Farid provided critical feedback for the overall manuscript. Senjian and Ferdous helped to develop the methodology from a machine learning perspective. Anita and Gary developed the discussion section and helped in interpreting the results from a marine scientist’s perspective. Robert assisted with the statistical analysis of the results and provided important revisions. All authors contributed critically to the drafts and gave final approval for publication.}

\funding{This research was partially supported by Australian Research Council Grants (DP150104251 and DE120102960)  and the Integrated Marine Observing System (IMOS) through the Department of Innovation, Industry, Science and Research (DIISR), National Collaborative Research Infrastructure Scheme.}

\acknowledgments{The authors acknowledge NVIDIA for providing a Titan-X GPU for the experiments involved in this research.}

\conflictsofinterest{The authors declare no conflict of interest.} 


\reftitle{References}


\externalbibliography{yes}
\bibliography{latest.bib}





\newpage
\appendix
\subsection{Class Distribution of Rottnest Island Data}

\begin{longtable}{llll}
\toprule
Label & Training Samples & Test Samples & CATAMI Class ID \\
\midrule
1     & 1      & 0     & AUC     \\
2     & 0      & 1     & AUS     \\
3     & 2      & 0     & BMC     \\
4     & 483    & 294   & BRYH    \\
5     & 20     & 13    & BRYS    \\
6     & 20     & 0     & CB      \\
7     & 1      & 0     & CBBF    \\
8     & 2      & 0     & CBBH    \\
9     & 7      & 0     & CBOT    \\
10    & 0      & 3     & CNHYC   \\
11    & 3      & 0     & CNHYD   \\
12    & 7      & 1     & CSBL    \\
13    & 44     & 19    & CSBR    \\
14    & 1      & 1     & CSBRBL  \\
15    & 15     & 3     & CSCOLBL \\
16    & 2      & 0     & CSCOR   \\
17    & 2      & 2     & CSCORBL \\
18    & 7      & 3     & CSDBL   \\
19    & 265    & 38    & CSE     \\
20    & 24     & 1     & CSEBL   \\
21    & 887    & 355   & CSF     \\
22    & 46     & 2     & CSFBL   \\
23    & 7      & 3     & CSM     \\
24    & 50     & 8     & CSSO    \\
25    & 1      & 0     & CSSOBL  \\
26    & 0      & 2     & CSST    \\
27    & 1      & 0     & CSSUBL  \\
28    & 1      & 1     & CST     \\
29    & 1      & 0     & CSTBL   \\
30    & 10     & 7     & EF      \\
31    & 47     & 2     & ESC     \\
32    & 15     & 1     & ESS     \\
33    & 102    & 31    & FELR    \\
34    & 0      & 3     & MAAG    \\
35    & 2644   & 2561  & MAAR    \\
36    & 37     & 0     & MACAU   \\
37    & 66     & 113   & MAECB   \\
38    & 1      & 1     & MAECG   \\
39    & 112762 & 43014 & MAECK (Kelp)  \\
40    & 2419   & 1124  & MAECR   \\
41    & 1733   & 173   & MAEFB   \\
42    & 1      & 1     & MAEFG   \\
43    & 2839   & 586   & MAEFR   \\
44    & 6744   & 1300  & MAENB   \\
45    & 29948  & 11686 & MAENR   \\
46    & 1252   & 2073  & MAFR    \\
47    & 2      & 0     & MAGB    \\
48    & 9      & 0     & MAGG    \\
49    & 1      & 0     & MAGR    \\
50    & 4      & 0     & MALAB   \\
51    & 2      & 0     & MALAR   \\
52    & 285    & 87    & MALCB   \\
53    & 3      & 1     & MAPAD   \\
54    & 1177   & 2391  & MASAR   \\
55    & 52     & 6     & MASB    \\
56    & 16571  & 3366  & MASCY   \\
57    & 137    & 0     & MASR    \\
58    & 24637  & 4846  & MATM    \\
59    & 2      & 0     & RH      \\
60    & 1505   & 163   & SC      \\
61    & 14     & 13    & SCC     \\
62    & 2      & 0     & SEAGSAA \\
63    & 18     & 3     & SEAGSAG \\
64    & 0      & 3     & SEAGSPA \\
65    & 1      & 3     & SEAGSPC \\
66    & 2      & 0     & SEAGSPS \\
67    & 1      & 0     & SEAGSZ  \\
68    & 106    & 15    & SHAD    \\
69    & 2013   & 1201  & SPC     \\
70    & 400    & 214   & SPCL    \\
71    & 110    & 125   & SPEB    \\
72    & 123    & 36    & SPEL    \\
73    & 289    & 347   & SPES    \\
74    & 69     & 0     & SPM     \\
75    & 23     & 6     & SUPBC   \\
76    & 164    & 4     & SUPBR   \\
77    & 9340   & 1893  & SUS     \\
78    & 68     & 1     & UNK    \\
\bottomrule 
\caption{Class Distribution of Rottnest Island Data}
\label{class}
\end{longtable}

\end{document}